\documentclass[conference]{IEEEtran}
\IEEEoverridecommandlockouts
\usepackage{cite}
\usepackage[utf8x]{inputenc} 
\usepackage{mathtools}
\usepackage{amsmath,amssymb,amsfonts}
\usepackage{algorithmic}
\usepackage{graphicx}
\usepackage{textcomp}
\usepackage{xcolor}
\def\BibTeX{{\rm B\kern-.05em{\sc i\kern-.025em b}\kern-.08em
    T\kern-.1667em\lower.7ex\hbox{E}\kern-.125emX}}
\usepackage{algorithm}
\usepackage[frozencache,cachedir=.]{minted}
\usepackage{multirow}
\usepackage{hyperref}
\hypersetup{
    colorlinks=true,
    linkcolor=blue}
\begin{document}

\title{A Novel Plug-in Module for Fine-Grained Visual Classification}

\author{\IEEEauthorblockN{Po-Yung Chou,
                          Cheng-Hung Lin~\IEEEmembership{Member,~IEEE}, and
                          Wen-Chung Kao~\IEEEmembership{Fellow,~IEEE}}
\IEEEauthorblockA{Department of Electrical Engineering, National Taiwan Normal University}
\thanks{Corresponding author: Cheng-Hung Lin (email: brucelin@ntnu.edu.tw).}}

\maketitle

\begin{abstract}
Visual classification can be divided into coarse-grained and fine-grained classification. Coarse-grained classification represents categories with a large degree of dissimilarity, such as the classification of cats and dogs, while fine-grained classification represents classifications with a large degree of similarity, such as cat species, bird species, and the makes or models of vehicles. Unlike coarse grained visual classification, fine-grained visual classification often requires professional experts to label data, which makes data more expensive. To meet this challenge, many approaches propose to automatically find the most discriminative regions and use local features to provide more precise features. These approaches only require image-level annotations, thereby reducing the cost of annotation. However, most of these methods require two- or multi-stage architectures and cannot be trained end-to-end. Therefore, we propose a novel plug-in module that can be integrated to many common backbones, including CNN-based or Transformer-based networks to provide strongly discriminative regions. The plugin module can output pixel-level feature maps and fuse filtered features to enhance fine-grained visual classification. Experimental results show that the proposed plugin module outperforms state-of-the-art approaches and significantly improves the accuracy to 92.77\% and 92.83\% on CUB200-2011 and NABirds, respectively. We have released our source code in Github \url{https://github.com/chou141253/FGVC-PIM.git}.

\end{abstract}


\section{Introduction}
\begin{figure}
\centerline{\includegraphics[width=8.5cm]{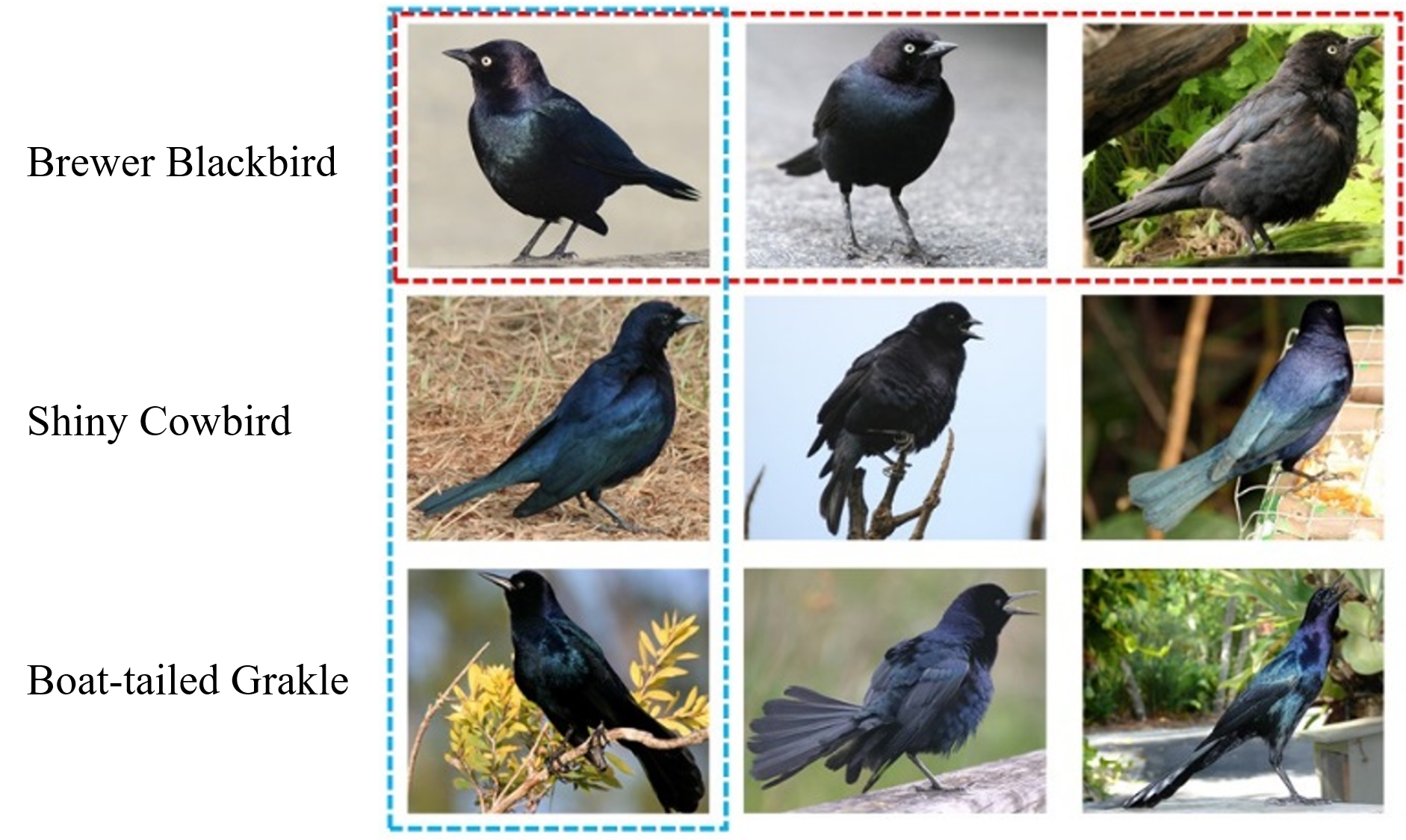}}
\caption{Three similar looking birds, Brewer Blackbird, Shiny Cowbird, and Boat-tailed Grakle. Each row represents three different appearances of the same bird. There are very few differences between different birds.}
\label{fig1}
\end{figure}

Visual classification can be divided into coarse-grained and fine-grained classification. Coarse-grained classification represents the classification of categories with a large degree of dissimilarity, such as the classification of cats and dogs, while fine-grained classification represents classifications with a large degree of similarity, such as bird species\cite{CUB_200_2011}\cite{NABirds}, dog species\cite{Stanford_Dogs}, and the makes or models of vehicles\cite{Stanford_Cars}. "Fine-grained" refers to more fine-grained divisions under common species classification. The challenges of fine-grained visual classification task is threefold. First, there is a lot of variation in the same category. Taking birds as an example, photos of the same bird from different angles can vary greatly in color and shape, as shown in Fig.\ref{fig1}. The second point is that objects of different subcategories may be very similar. As shown in Fig.\ref{fig1}, the textures of the three birds are very similar. The third point is that unlike coarse grained classification, fine-grained classification often requires professional experts to label data, which makes data more expensive. For the above reasons, frameworks that perform very well on coarse-grained classification tasks such as ResNet\cite{ResNet}, EfficientNet\cite{EfficientNet}, Vision Transformer (ViT) \cite{ViT} will be very limited for fine-grained classification.

To find strong discriminative regions to enhance the fine-grained visual classification, the proposed approaches can be classified into three categories. The first type of method finds regions through Region Proposal Network (RPN)\cite{Faster_RCNN}, such as NTS-Net\cite{NTS_Net}, FDL\cite{FDL}, and StackedLSTM\cite{Stacked_LSTM}. The second type of method strengthens the feature map through the attention mechanism, such as CAL\cite{CAL}, MA-CNN\cite{MA_CNN}, MAMC\cite{MAMC}, API-Net\cite{API_Net}, and WS-DAN\cite{WS_DAN}. The third type of method uses the strength of the attention-map in the self-attention mechanism as the judgment of discriminative regions, such as TransFG\cite{TransFG} and FFVT\cite{FFVT}. The first two approaches are mainly based on convolutional neural networks (CNN), such as ResNet\cite{ResNet}, DenseNet\cite{DenseNet}, and EfficientNet\cite{EfficientNet}, while the third is based on ViT to implement fine-grained visual classification.

After finding these regions, the above approaches re-input the original image and feature maps to the network by cropping and resizing, or use an attention mechanism to strengthen the relationship between feature maps. The disadvantage of these approaches is that most of them require two-stage or multi-stage complex architectures and cannot be trained end-to-end. Furthermore, these localization methods tend to generate large prediction regions and are less able to focus on subtle local features. On the other hand, the Vit-based method directly uses the attention-map in the self-attention mechanism as the basis for selecting regions, avoiding the feedback-based architecture and achieving efficient end-to-end training. But this method is difficult to generalize to convolutional neural networks or other architectures, which means that its scalability is limited. 

In order to better understand the relationship between feature maps and object positions, we explore the relationship between the object detection models and the FGVC methods. From Faster-RCNN\cite{Faster_RCNN}, YOLO\cite{YOLO}\cite{YOLO_v4}, and RetinaNet\cite{RetinaNet}, we can find that feature maps can be rich in location and category information. Segmentation methods such as Mask-RCNN\cite{Mask_RCNN} give pixel-level predictions, i.e. fine-grained predictions. Although these methods demonstrate that feature maps are rich in semantics, they are learned by relying on human-annotated region information, unlike the FGVC task, which only uses image-level annotation  for training. Therefore, we discuss Weakly Supervised Object Detection(WSOD) methods, such as WSDDN\cite{WSDDN}, OICR\cite{OICR}, and WCCN\cite{WCCN}, etc. The concept of the WSOD method is that the response of feature maps reflects the location of objects, and we can use these feature maps to complete bounding box prediction through an weakly supervised multi-stage architecture and loss function design.

Based on the above analysis, we propose a plug-in module that can be added to many common backbones, which can be implemented on CNN-based or Transformer-based architectures. In addition, the plugin module outputs pixel-level feature maps and fuses the filtered features.

To better detect key features of various sizes, we add the Feature Pyramid Network (FPN)\cite{FPN} to the backbone network to mix spatial features of different scales. This architecture is very common in object detection tasks, and this approach can improve the quality of local representation. 

\begin{figure}
\centerline{\includegraphics[width=8.5cm]{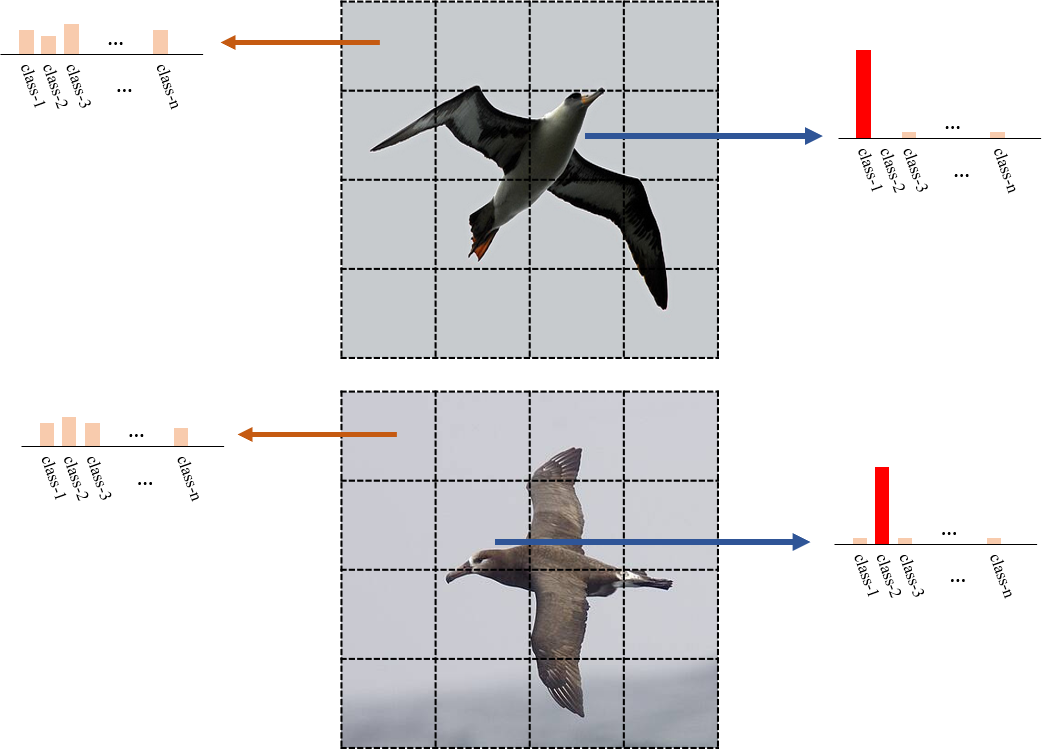}}
\caption{The probability distribution of the predicted value in different patches. The probability distribution of background patches will be flat, while the probability distribution of some objects will be shaped.}
\label{fig2}
\end{figure}

The idea of the plugin module is composed of three operations, including division, competition, and combination. Consider the example in Fig.\ref{fig2}, the patches obtained by dividing the background are almost solid colours. This kind of patch will appear widely in different categories of data. If we use these patches for training, we can expect that the predicted distribution of the test images will be very flat. On the contrary, if we use the patches which contain discriminative objects for training, the predicted results would be more distinguishable.

The reason is that nearly identical training data but labeled as different classes can lead to difficulty in converging on the sample space during the backward propagation. This phenomenon gives us the idea that the predicted class score can be used to divide the feature map into object candidate regions and background candidate regions. Divided object features of different scales will be fused for classification prediction, while the correction of the background area is aimed at a flat probability distribution. In this way, we can remove background noise and focus on important regions.

This paper has the following two main contributions.

\begin{itemize}
\item We propose a novel plug-in network that can be applied to various models. This network integrates novel background segmentation  and feature fusion techniques, which can effectively improve the accuracy of fine-grained visual classification.
\item The proposed network outperforms state-of-the-art networks and improves the accuracy to 92.77\% (+0.97\%) and 92.83\% (+1.83\%) on CUB200-2011\cite{CUB_200_2011} and NABirds\cite{NABirds}, respectively.
\end{itemize}

\section{Related Work}
In this section, we will introduce the proposed architecture for coarse-grained image recognition tasks, which are mainly divided into convolution methods and self-attention mechanisms. And then, we will introduce the object detection task and discuss the ideas we will use. Finally, we will analyze several fine-grained visual classification models that have performed well recently, and compare these methods in experimental results.

\subsection{Backbones}
CNN has been quite successful in image recognition. Since the accuracy of AlexNet\cite{AlexNet} on ImageNet\cite{ImageNet} has greatly surpassed traditional methods, various convolutional architectures have been proposed to improve the accuracy of recognition. For example, ResNet\cite{ResNet} and DenseNet\cite{DenseNet} have performed quite well on various recognition tasks using deep network structures with shortcut connections, and their pre-trained models on ImageNet have also been successfully transferred to various computer vision tasks. EfficientNet\cite{EfficientNet} further discusses the depth, width and input resolution of convolutional networks to optimize the network structure. The method successfully maintains high accuracy with a small number of parameters.
Since the publication of Transformer\cite{Transformer}, the self-attention mechanism has been widely used in the field of natural language processing and computer vision, and its performance is quite good. For example, Vit\cite{ViT} achieved 90.7\% accuracy on ImageNet, which convert the patch of the image into a token through linear transformation, and complete the global information exchange through the self-attention mechanism. However, its disadvantage is that there is no hierarchical expression ability for the features of local regions. In SwinTransformer\cite{Swin_T}, the extraction of local region features at different scales is done through a multi-layer self-attention structure. Because of the success of the above architectures, we will use these models as the backbone network to test the capabilities of the proposed modules.

\begin{figure}
\centerline{\includegraphics[width=8.5cm]{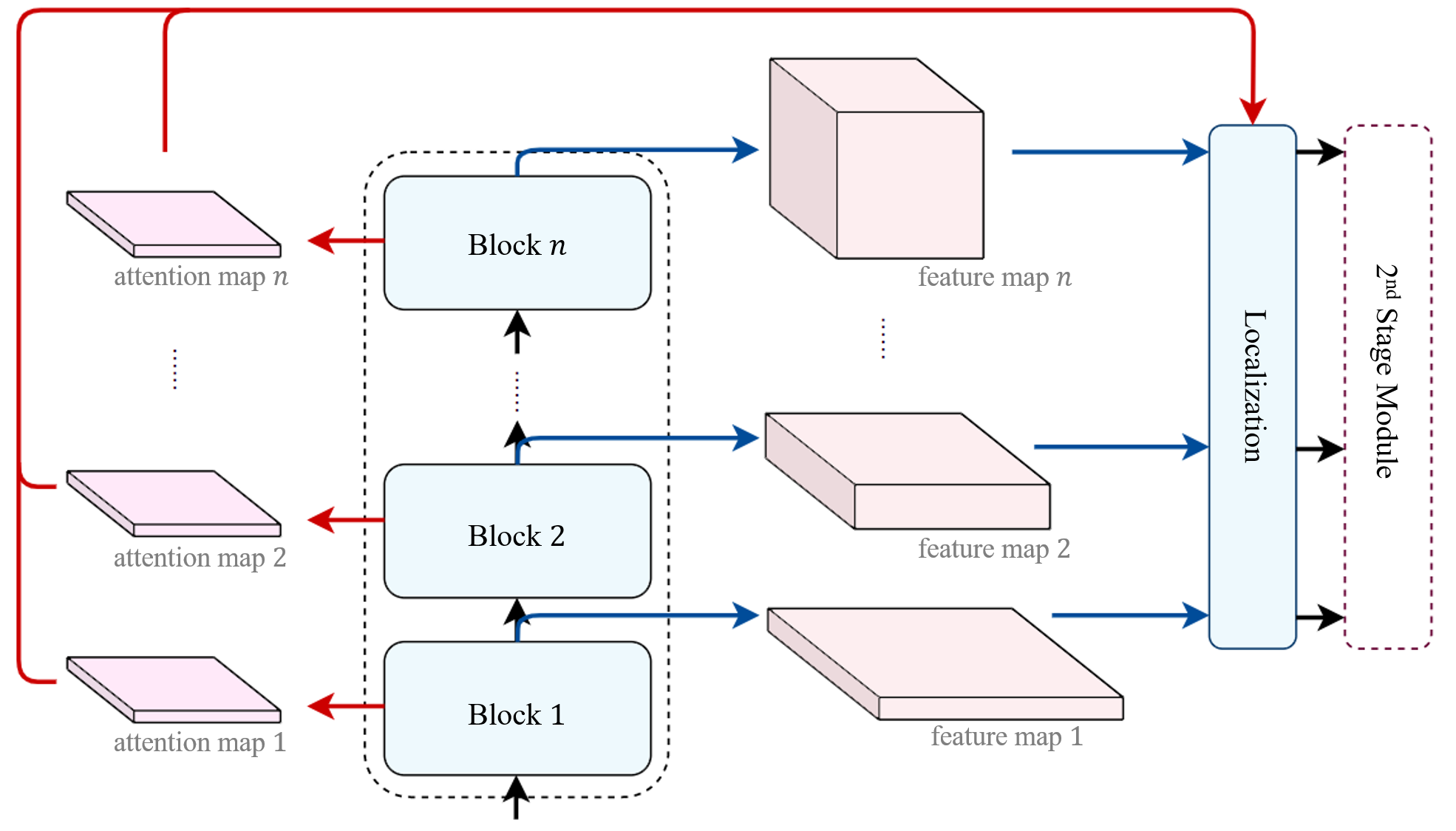}}
\caption{Schematic flow of the past fine-grained visual classification networks. The model architecture and training process of these methods are summarized in Fig.\ref{fig3}. The light blue square ($ block_{1}\sim block_{n}$) in the middle represents the blocks of the main backbone network (e.g. ResNet\cite{ResNet}, ViT\cite{ViT}). The transfer direction is shown by the arrow. The light red solid square on the right side of the backbone network represents the feature map of the input image after passing through the convolutional network. This feature map can be used to complete object positioning or re-cutting. We call this process the localization module. Finally, the integrated information will be sent to the second-stage model, which is represented by a dotted square. The second-stage model may be a specially designed structure, or it may be the original network. In this way, local features can be used more accurately to complete the identification. The light purple solid square on the left side of the backbone network represents the attention-map generated by the Transformer, and the subsequent process is the same as the previous process based on convolution-based models.}
\label{fig3}
\end{figure}

\subsection{Object detection}
The goal of object detection is to find the location and classification of objects, so the overall architecture and idea will be very similar to the fine-grained visual classification task. The difference is that the purpose of the fine-grained visual classification is not whether the regions of the object are found, but whether the regions with distinguishable features are found, and whether these features can be more effectively used for identification.

First of all, we explore the supervised object detection networks. For example, Faster-RCNN\cite{Faster_RCNN} predicts whether each pixel position on the feature map is an object through Region Proposal Network (RPN), and then predicts the category of the object region. YOLO\cite{YOLO}\cite{YOLO_v4} and RetinaNet\cite{RetinaNet} complete the prediction of position and category through an overall network. The above methods all learn the region of the object from the manual annotation, and complete the identification for the region of the object.

Let us explain this process in another angle. In the training phase, the network needs to pay attention to these regions with objects, and then complete the classification prediction of the region. The whole process is very close to the method proposed by FGVC, but the FGVC method we intend to propose does not use the target location information as training data. Instead, we extend to weakly supervised learning methods in object detection that do not rely on manually annotating object regions.

\subsection{Weakly supervised object detection}\label{AA}
Weakly Supervised Object Detection (WSOD) relies on meaningful feature maps from deep neural networks. B. Zhou et al.\cite{scenecnn_iclr15} observed that by learning object class labels, the representation of objects in space can be learned, i.e., virtual labeling of object locations can be done through information on feature maps. For example, WSDDN\cite{WSDDN} uses the pre-trained model as the feature extractor, and completes the prediction of the features of the candidate regions through a two-stream network, and then uses the prediction result to filter the regions. That is, the identification is completed by the quality of the prediction of each region.

OICR\cite{OICR} continues the former method and adds a multi-stage classifier to complete more accurate positioning. WCCN\cite{WCCN} completes the division of candidate regions through Class Activation Map (CAM)\cite{CAM} at first, the principle of CAM is that map prediction score on the previous features map to generate the class-specific heat map, and then use the second-stage model to screen better candidate regions. Methods such as ACoL\cite{ACoL} and SPG\cite{SPG} are also based on CAM to complete localization. DANet\cite{DA_Net} learns more diverse features by comparing the fusion category and uses CAM to locate more complete objects. CASD\cite{CASD} trains OICR\cite{OICR} by means of self-distillation, and MIST\cite{MIST} perform self-training by generating virtual labels. The above methods show that category labels can provide features rich in object location and locate more class-discriminating regions. However, the task with WSOD is still not the same as with FGVC. The goal of the FGVC task is no longer to detect complete objects, but to find the most critical regions and exploit these regions for better recognition. 

\subsection{Fine-grained visual classification}
The main purpose of FGVC is to locate highly discriminatory regions. Many methods such as NTS-Net\cite{NTS_Net}, FDL\cite{FDL}, StackedLSTM\cite{Stacked_LSTM} find strong discriminative preselected boxes through Region Proposal Network (RPN), then resize the preselected boxes to a fixed size, and use these local strong features to identify, that is to say, these methods will actually divide the identification task into two stages.

Mix+\cite{Mix_plus} uses the attention map to find the regions with strong discrimination and uses this map to complete the mixed reinforcement. CCFR\cite{CCFR} recognizes the global features through these regional features, and learn better local region features by triplet loss and scale-separated NMS. 

Attention mechanisms are also widely used in FGVC tasks, such as MACN\cite{MA_CNN}, WS-DAN\cite{WS_DAN} and CAL\cite{CAL} to learn the location of objects, and extract the characteristics of a specific location in a feature map through the attention-map. In addition to being used to learn location information, the attention mechanism can also improve the expressiveness of features. For example, MAMC\cite{MAMC} and API-Net\cite{API_Net} learn unique features by simultaneously training the differences between a pair of similar images through the attention mechanism, and CAP\cite{CAP} fuses different local features through the attention mechanism. The above methods all show that the attention mechanism is a very powerful method.

Since the release of Vision Transformer(ViT)\cite{ViT}, the ViT architecture has performed quite well in the field of image recognition, so many approaches are proposed to apply this architecture to the FGVC task. For example, FFVT\cite{FFVT}, AFTrans\cite{AFTrans}, RAMS-Trans\cite{RAMS_Trans} and TransFG\cite{TransFG} use ViT as the backbone network, and use the attention-map generated by the image in self-attention as the search for strong discriminative regions. Finally, the features of these regions are processed to complete the recognition task. This approach is similar to the aforementioned method based on the convolutional network. As shown in the left part of Fig.\ref{fig3}, the difference is that the strength of the attention map is used instead of the response of the feature map.

\begin{algorithm}[t]
\caption{Weakly Supervised Selector, PyTorch-like Code} 
\vspace{10pt}
\begin{minted}[fontsize=\footnotesize]{python}

import torch
import torch.nn as nn

class WSS(nn.Module):
  
    def __init__(self, 
               in_channel: int, 
               num_classes: int, 
               num_selects: int):
        super().__init__()
        self.fc = nn.Linear(in_channel, num_classes)
        self.num_selects = num_selects
    
    def forward(self, x):
        # [B, H×W, C] = x.shape
        # return class_prediction, selected_features
        logits = torch.softmax(self.fc(x))
        _, ids = \
            torch.sort(logits, -1, descending=True)
        selection  = ids[:, :self.num_selects]
        return logits, torch.gather(x, 1, selection)

\end{minted}
\vspace{10pt}
\end{algorithm}

\section{A NOVEL PLUG-IN  MODULE FOR FINE-GRAINED VISUAL CLASSIFICATION}

To find strong discriminative regions for fine-grained classification tasks, we propose a plug-in module that can be applied to mainstream backbone networks such as ResNet\cite{ResNet}, EfficientNet\cite{EfficientNet}, and ViT\cite{ViT}. The overall design concept is to treat each pixel (or patch) on the feature map as an independent feature, which can represent its region. Then we classify these features and use the classification ability as a basis for distinguishing. Finally, the entire network can be completed through end-to-end training. In this chapter, the design of the module structure, the use of the loss function, and the combination with each framework will be introduced in detail.

\begin{figure}[!tp]
    \centerline{\includegraphics[width=8.5cm]{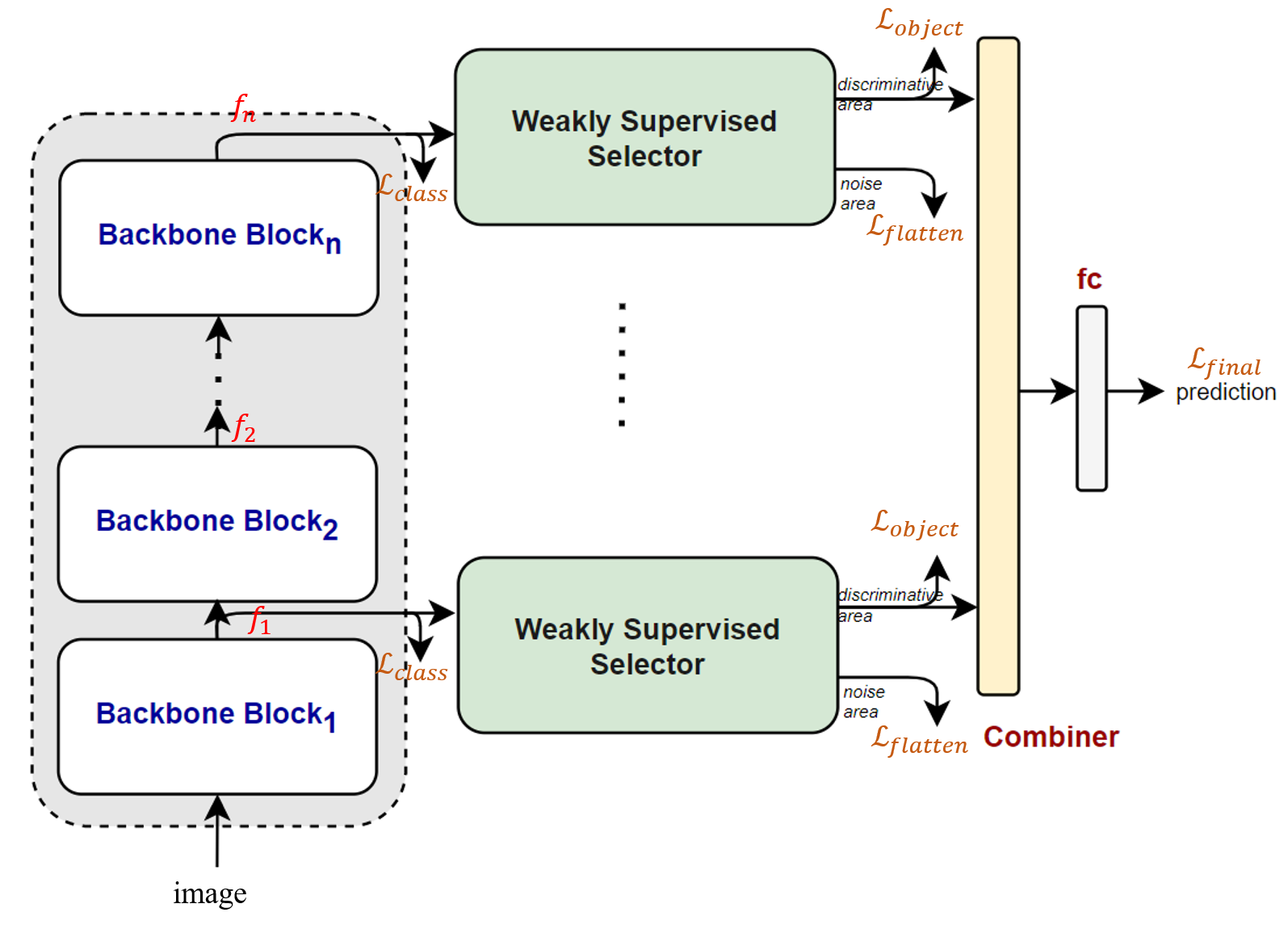}}
    \caption{Schematic flow of the proposed plug-in module. Backbone Blockk represents the kth block in the backbone network. When the image is input to the network, the feature map output by each block will be input into the Weakly Supervised Selector to screen out areas with strong discrimination or areas that are less related to classification. Finally, a Combiner is used to fuse the features of the selected results to obtain the prediction results. The Lfinal represents the loss function.}
    \label{fig4}
\end{figure}

\subsection{Module design}
The common point of the FGVC and WSOD methods discussed above is to find a region with strong discrimination. FGVC cuts this area or increases attention during secondary training process, while WSOD frameworks often use multiple instance learning (MIL) for more accurate localization. These two tasks are still slightly different in positioning targets, but these architectures can prove that the pixel-level features in the feature map can represent the importance of this area in the classification task. In the following, we call the value in the pixel of the feature map as the feature point. The dimension of this feature point is $R^{C}$, where $C$ represents the output feature dimension size of this block.

We adopt a very simple design, which passes each feature point through a fully connected layer to predict the category. When the highest probability of the predicted result after softmax is greater than a certain value, the feature point is considered as a helpful feature and will be reserved for subsequent fusion. Conversely, the feature point is considered less helpful for fine-grained classification.

Based on the above concepts, we design an architecture that can be trained end-to-end, as shown in Fig.\ref{fig4}. The area framed by the dotted line is the backbone model, and we use $f_{i} \in R^{C\times H\times W}$ to represent the feature map output by the $i^{th}$ block in the backbone network, where $H$ represents the height of the feature map, $W$ represents the width of the feature map, $C$ represents the size of the feature dimension. This feature map is then fed into a weakly supervised selector and each feature point will be classified by a linear classifier. The feature map after this step is denoted as $f_{i} \in R^{C\prime\times H\times W}$, where $C\prime$ is equal to the number of target classes. Then, the class prediction probability of each feature point is obtained through softmax, and the first few feature points with high confidence score will be selected in the weakly supervised selector.

The selection algorithm is shown in \textbf{Algorithm 1}, and the selected features are fused through the fusion model, whose architecture is shown in Fig.\ref{fig4}. In order to complete the feature fusion, we designed two different architectures. The first one is implemented through a fully connected layer. Assume that the total number of selected feature points is $N$. Before the feature maps are input to the fully connected layer, they are first concatenated together for the feature dimension. Therefore, the feature map with dimension $ R^{N \times C} $  produces a prediction result with dimension $R^{C\prime}$ after passing through the fully connected layer. This architecture can recombine the selected local features into global features that can represent the entire image. The second architecture is implemented through graph convolution, which treats all selected feature points as a graph structure, where nodes represent features at different spatial locations and scales. The graph is input into the graph convolutional network, which can learn the relationship between different nodes. And then, the feature points are aggregated into several super nodes through the pooling layer, and finally the features of these super nodes are averaged, and a linear classifier is used to complete the prediction. The advantage of this approach is that the features of each point can be integrated more efficiently without corrupting the results output by the backbone model. Therefore, we finally use graph convolution as the feature fusion mechanism.

In addition, in order to allow the model to extract the features of small regions more effectively, we add FPN to the backbone network to effectively fuse features of different scales to achieve more accurate recognition results.

\subsection{The process of forward and backward propagation}

\begin{figure}[!tp]
\centerline{\includegraphics[width=8.5cm]{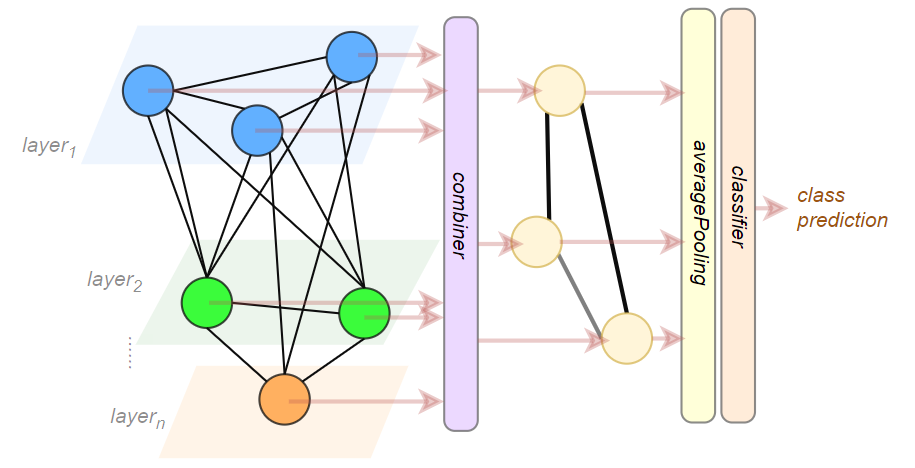}}
\caption{The schematic diagram of the Combiner. The input is the selected feature points of different scales, which are represented by different colors. The features are fused through the Combiner, and finally the recognition is completed through these fused features.}
\label{fig5}
\end{figure}


The goal of this architecture is to perform fine-grined classification, so it does not use any artificial labels as training targets other than image-level annotations. The first training goal is to make the features of each feature map $f_{i}$ have the ability to classify. In order to calculate the overall loss, we first average the prediction output of all feature points, as shown in the following Eq.(\ref{eq1}), where $f_{l,s} \in R^{C\prime}$ represents the feature point at position $s$ of the $l^{th}$ block of feature map, $S$ denotes the output feature map space of this block, and the size is $H\times W$. Then the class loss of the entire block is calculated through Cross Entropy, as shown in Eq.(\ref{eq2}).

\begin{equation} \label{eq1}
    z_{l} = \frac{1}{H \times W} \sum_{s \in S} f_{l,s}
\end{equation}

\begin{equation} \label{eq2}
    L_{b} = - \sum_{l \in L} log(z^{cls} _{l})
\end{equation}

In addition to training the overall classification, the selected position of the feature map is marked by $Mask \in R^{H \times W}$. This $Mask$ is a binary data type, where 1 represents the selected area and 0 represents the dropped area. Therefore, $S \odot (Mask)$ represents feature points with strong discrimination, while $S \odot (\sim Mask)$ represents dropped feature points. Eq. (\ref{eq3}) represents the sum of the features of the selected region in the $l^{th}$ block. The loss function of the selected category is $L_s$ defined as Eq.(\ref{eq4}). Eq. (\ref{eq5}) represents the sum of the features that were not selected (or can be said to be selected as background). The flattening loss function for the dropped area is $L_n$ defined as Eq.(\ref{eq6}). 

\begin{equation} \label{eq3}
    h_{l} = \sum_{s \in S \odot(Mask)} f_{l,s}
\end{equation}

\begin{equation} \label{eq4}
    L_{s} = - \sum_{l \in L} \sum_{i \in C} log(h^{i}_{l})
\end{equation}

\begin{equation} \label{eq5}
    n_{l} = \sum_{s \in S \odot(\sim Mask)} f_{l,s}
\end{equation}

\begin{equation} \label{eq6}
    L_{s} = - \sum_{l \in L} \sum_{i \in C'} log(1 - n^{i}_{l})
\end{equation}

The flattened output is designed to express that this area is less helpful for classification. In fact, this approach is like predicting the "score" of the foreground or background in the object detection framework. In this paper, the highest probability value of the softmax is taken as this score.

Then, the selected feature $f_{s} \in R^{N\times C}$ will be input into the Combiner to generate prediction results of mixed scales, the output feature is $f_{comb} \in R^{N\prime \times C}$,  where $N\prime$ is the number of super nodes after condesing input nodes. Finally, the features on these super nodes are averaged and input into a linear classifier to output predictions. The Combiner category prediction loss is calculated through Cross Entropy, and this loss function is represented by $L_{c}$. The entire loss function is defined as Eq.(\ref{eq7}). This loss function is the weighted sum of the above loss functions, where $\lambda_{b}$, $\lambda_{s}$, $\lambda_{n}$, and $\lambda_{c}$ are the weights of $L_{b}$, $L_{s}$, $L_{n}$, and $L_{c}$, respectively:

\begin{equation} \label{eq7}
    L = \lambda _{b} L_{b} + \lambda _{s} L_{s} + \lambda _{n} L_{n} + \lambda _{c} L_{c}
\end{equation}

In fact, during training phrase, we set $L_{b}$=1, $L_{s}$=0, $L_{n}$=5, $L_{s}$=1, and do not use the selected category loss function $L_s$ as the training target because the prediction loss $L_{c}$ through the Combiner category already has the same function.

The overall training goal is to locate a local area with strong discrimination, and improve the recognition result through the features of this area. This goal is very similar to the functionality of the previous WSOD and FGVC frameworks, except that finding the local locations (or positions of objects) with strong discrimination are mostly done by the responses of feature maps, often called heatmaps. The downside of this approach is that many algorithms are needed to find the heatmap, and many architectures require a two-stage approach to revise the model.

In this paper, we propose a simple and easy-to-implement method that mainly uses local features to predict classes. This makes it difficult to "distinguish" the predicted values of local background regions or locally similar parts of different classes. In this case, a strong selection criterion -- maximum predicted probability. Finally, the selected local features are fused into global features to complete the final prediction. This method can be easily applied to various mainstream backbone network and only one stage of end-to-end training is required.
 
Contrastive learning is widely used in FGVC tasks. For example, TransFG\cite{TransFG} learns better features through contrastive learning of features. API-Net\cite{API_Net} uses ranking loss to learn the difference between a pair of image features. CCFR\cite{CCFR} uses triplet loss to learn regions of discrimination. Contrastive learning seems to play an important role for the FGVC model. However, training a model through contrastive learning usually requires additional parameters, such as margin distance or temperature, etc. Because we hope to keep the training as simple as possible, contrastive learning is not used in this design, only Cross Entropy is used.

\begin{table}[!tp]
\caption{Comparison of different methods on CUB-200-2011.}
\begin{center}
\begin{tabular}{c|c|c}
Method & Backbone & Accuracy(\%) \\
\hline
ResNet-50\cite{TransFG} & ResNet-50 & 84.5 \\
ResNet-50(Ours) & ResNet-50 & 88.2 \\
API-Net\cite{API_Net} & DenseNet-161 & 90.0 \\
Mix+\cite{Mix_plus} & ResNet-50 & 90.2 \\
StackedLSTM\cite{Stacked_LSTM} & GoogleNet & 90.4 \\
CAL\cite{CAL} & ResNet-101 & 90.6 \\
DeepFVE\cite{DeepFVE} & InceptionV3\cite{Inception_v3} & 91.0 \\
CCFR\cite{CCFR} & ResNet-50 & 91.1 \\
RAMS-Trans\cite{RAMS_Trans} & ViT-B\_16 & 91.3 \\
TPSKG\cite{TPSKG} & ViT-B\_16 & 91.3 \\
AFTrans\cite{AFTrans} & ViT-B\_16 & 91.5 \\
FFVT\cite{FFVT} & ViT-B\_16 & 91.6 \\
TransFG\cite{TransFG} & ViT-B\_16 & 91.7 \\
CAP\cite{CAP} & Xception\cite{Xception} & 91.8 \\
\hline
PIM $^{\mathrm{*}}$ (ours) & Swin-T & \textbf{92.8}  \\
\hline
\multicolumn{3}{l}{ $^{\mathrm{*}}$ PIM denotes the proposed plug-in module.}

\end{tabular}
\label{tab1}
\end{center}
\end{table}

\section{Experimental Results}

\begin{table}
\caption{Comparison of different methods on NB-Birds.}
\begin{center}
\begin{tabular}{c|c|c}
Method & Backbone & Accuracy(\%) \\
\hline
Cross-X\cite{Cross_X} & ResNet-50 & 86.4 \\
PAIRS\cite{PAIRS} & ResNet-50 & 87.9 \\
API-Net\cite{API_Net} & DenseNet-161 & 88.1 \\
CS-Part\cite{CS_PART} & ResNet-50 & 88.5 \\
MGE-CNN\cite{MGE_CNN} & ResNet-101 & 88.6 \\
FixSENet-154\cite{FixSeNet_154} & SENet-154 & 89.2 \\
TransFG\cite{TransFG} & ViT & 90.8 \\
CAP\cite{CAP} & Xception & 91.0 \\
\hline
PIM$^{\mathrm{*}}$(ours) & Swin-T & \textbf{92.8} \\
\hline
\multicolumn{3}{l}{$^{\mathrm{*}}$PIM denotes the proposed plug-in module.}

\end{tabular}
\label{tab2}
\end{center}
\end{table}

\begin{table}[!tp]
\caption{ Improvement of different backbones with our additional structure.}
\begin{center}
\begin{tabular}{|c|c|c|}
\hline
Backbone & Original & +PIM \\
\hline
ResNet-50 & 88.2\% & 89.5\%(+1.3\%) \\
\hline
EfficientNet-7B$^{\mathrm{*}}$ & 88.2\% & 89.9\%(+1.7\%) \\
\hline
Vit & 90.1\% & 91.0\%(+0.9\%) \\
\hline
Swin-T & 91.9\% & 92.8\%(+0.9\%) \\
\hline
\multicolumn{3}{p{26em}}{$^{\mathrm{*}}$The pre-training dataset for EfficientNet-7B is ImageNet-1K, while the others are ImageNet-21K or -22K.}

\end{tabular}
\label{tab3}
\end{center}
\end{table}

\begin{table}[!tp]
\caption{Comparision of different evaluation methods.}
\begin{center}
\begin{tabular}{|c|c|}
\hline
Evaluation methods & Accuracy(\%) \\
\hline
block 1 average score & 91.70 \\
\hline
block 2 average score & 91.75 \\
\hline
block 3 average score & 92.18 \\
\hline
block 4 average score & 92.16 \\
\hline
combiner score & 91.80 \\
\hline
top-1 average score  & 91.62 \\
\hline
top-2 average score  & 91.99 \\
\hline
top-3 average score  & 92.15 \\
\hline
top-4 average score & 92.75 \\
\hline
top-5 average score  & \textbf{92.77} \\
\hline
\end{tabular}
\label{tab4}
\end{center}
\end{table}

In this section, we will introduce the datasets and the setting of experimental hyperparameters, then compare with some current state-of-the-art methods, and finally discuss some factors that affect the recognition accuracy and visualize the results.

\subsection{Datasets and implementation details}

The datasets we use are CUB200-2011\cite{CUB_200_2011} and NA-Birds\cite{NABirds} two fine-grained bird identification datasets. The CUB200-2011 dataset has a total of 200 bird categories, including 5,994 training images and 5,794 testing data. Each category contains about 30 training data. NA-Birds has 555 bird species, 23,929 training images and 24,633 test images. Both datasets provide image-level annotations and keypoint locations, but only  image-level annotations  will be used in this paper.

When ResNet-50\cite{ResNet}, EfficientNet-B7\cite{EfficientNet} and ViT\cite{ViT} are adopted as backbone networks, the input image is a 448×448 color image. When  Swin-T\cite{Swin_T} is used as the backbone network, the input image is a 384×384 color image. 

The methods of data augmentation is as follows. If the input image size is 384×384, the first step is to scale the image to 510×510, and if the input image size is 448×448, it is scaled to 600×600. In training phrase, data augmentation is performed via Randon Crop, Random HorizontalFlip, and Random GaussianBlur while in testing phrase, Center Crop is used. During training, the learning rate is set to 0.0005, and the cosine decay is used; the weight decay is set to 0.0005; SGD is used as the optimizer, and the batch size is set to 8; a total of 50 epochs are trained.

All experiments are completed on a single Nvidia GeForce RTX 3090, and the Pytorch toolbox is used as the main implementation substrate. It takes about 3 hours to complete the training on CUB200-2011, and about 15 hours for NA-Birds.

\begin{table}[!tp]
\caption{Comparision of different evaluation methods.}
\begin{center}
\begin{tabular}{|c|c|}
\hline
Selections Number & Combiner-1(\%) \\
\hline
[32,32,32,32] & 92.61 \\
\hline
[256,128,64,32] & 92.67 \\
\hline
[512,256,128,64] & 92.37 \\
\hline
[1024,512,128,64] & 92.48 \\
\hline
[1024,512,128,128] & 92.54 \\
\hline
[2048,512,128, 32] & 92.77 \\
\hline
[2048,512,128,128] & 92.34 \\
\hline
[2304,576,144,144] & 92.34 \\
\hline
\end{tabular}
\label{tab5}
\end{center}
\end{table}

\begin{table}[!tp]
\caption{Different selection number of each layer and its corresponding accuracy..}
\begin{center}
\begin{tabular}{|p{5em}|p{3em}|p{6em}|p{6em}|}
\hline
\hline
Selections & Block id & Selected Accuracy(\%) & Dropped   Accuracy(\%) \\
\hline
\hline
\multirow{4}{5em}{[256, 128, 64, 32]} &  1 & 91.68 & 81.43 \\ \cline{2-4} 
                       &  2 & 91.62 & 71.03 \\ \cline{2-4} 
                       &  3 & 92.17 & 50.37 \\ \cline{2-4} 
                       &  4 & 92.29 & 70.98 \\
\hline
\hline
\multirow{4}{5em}{[2048, 512, 128, 32]} &  1 & 91.82 & 10.01 \\ \cline{2-4} 
                       &  2 & 91.44 & 10.08 \\ \cline{2-4} 
                       &  3 & 92.68 & 9.12 \\ \cline{2-4} 
                       &  4 & 92.72 & 8.94 \\
\hline
\hline
\end{tabular}
\label{tab6}
\end{center}
\end{table}

\subsection{Compare with state-of-the-art approaches}

In Table \ref{tab1}, we compare our plug-in module (PIM) with state-of-the-art methods on CUB200-2011 dataset. We use a Swin-T model pre-trained on ImageNet22K as the backbone. Table \ref{tab1} show that the proposed PIM  can reach 92.8\% in Top-1 accuracy, which is 1.0\% higher than the previous best method. Table \ref{tab2} shows that the proposed PIM can reach 92.8\% in Top-1 accuracy on NA-Birds dataset, which is 1.8\% higher than the previous best method. This results show that the proposed PIM can effectively fuse features of different scales, and this feature can effectively identify fine-grained categories.

To better discuss the accuracy improvement brought by this approach, we test the proposed PIM on four mainstream backbones. As shown in Table \ref{tab3}, the four mainstream backbones, including ResNet-50\cite{ResNet}, EfficientNet\cite{EfficientNet}, Vit\cite{ViT}, and Swin-T\cite{Swin_T} can be improved after adding our proposed PIM. The accuracy rate of Swin-T before joining PIM has exceeded state-of-the-art approaches, which shows that when the capacity of the backbone is good enough, the performance in downstream tasks will be quite good. So it can be predicted that these fine-grained recognition tasks will also become easier to complete after pre-training on large datasets through self-supervised methods. In this experiments, the pre-training dataset for EfficientNet-7B is ImageNet-1K, while the others are ImageNet-21K or -22K.

\begin{table}[!tp]
\caption{Comparision of different combiner structure.}
\begin{center}
\begin{tabular}{|c|c|c|}
\hline
Type & Number of Layers & Accuracy(\%) \\ \hline
ADD & 0 & 92.11 \\ \hline
MLP & 1 & 92.06 \\ \hline
\multirow{3}{*}{GCN} & 1 & 92.77 \\ \cline{2-3} 
                       & 2 & 92.58 \\ \cline{2-3} 
                       & 3 & 92.67 \\ 
\hline
\end{tabular}
\label{tab7}
\end{center}
\end{table}

\begin{table}[!tp]
\caption{Different number of GCN layers and nodes.}
\begin{center}
\begin{tabular}{|c|c|c|}
\hline
Layers & Numbers(\# of output nodes / \# of input nodes) & Accuracy(\%) \\ 
\hline
\multirow{8}{*}{1} & 1/128 & 92.37 \\ \cline{2-3}
                   & 1/64 & 92.51 \\ \cline{2-3}
                   & 1/32 & \textbf{92.77}  \\ \cline{2-3}
                   & 1/16 & 92.48 \\ \cline{2-3}
                   & 1/8 & 92.56 \\ \cline{2-3}
                   & 1/4 & 92.49 \\ \cline{2-3}
                   & 1/2 & 92.48 \\ \cline{2-3}
                   & 1/1 & 92.30 \\ 
\hline
\multirow{4}{*}{2} & 1/2, 1/4 & 92.15 \\ \cline{2-3}
                    & 1/4, 1/8 & 92.32 \\ \cline{2-3}
                    & 1/8, 1/16 & 92.56 \\ \cline{2-3}
                    & 1/16, 1/32 & 92.44 \\ \cline{2-3}
                    & 1/32, 1/64 & 92.61 \\ \cline{2-3}
                    & 1/64, 1/128 & 92.68 \\ \cline{2-3}
                    & 1/128, 1/256 & 92.63 \\
\hline
\multirow{3}{*}{3} & /2, /4, /8 & 91.63 \\ \cline{2-3}
                   & /4, /8, /16 & 92.11 \\ \cline{2-3}
                   & /8, /16, /32 & 92.25 \\ \cline{2-3}
                   & /16, /32, /64 & 92.48 \\ \cline{2-3}
                   & /32, /64, /128 & 92.53 \\
\hline

\end{tabular}
\label{tab8}
\end{center}
\end{table}

\begin{table}[!tp]
\caption{ Improvement of different backbones with our additional structure.}
\begin{center}
\begin{tabular}{|c|c|}
\hline
Components & Accuracy(\%) \\
\hline
Backbone & 91.9 \\ \hline
Backbone + FPN & 92.0 \\ \hline
Backbone + FPN + Selector & 92.1 \\ \hline
Backbone + FPN + Selector + Combiner & 92.8 \\ \hline
\end{tabular}
\label{tab9}
\end{center}
\end{table}

\begin{figure*}
\centerline{\includegraphics[width=16.8cm]{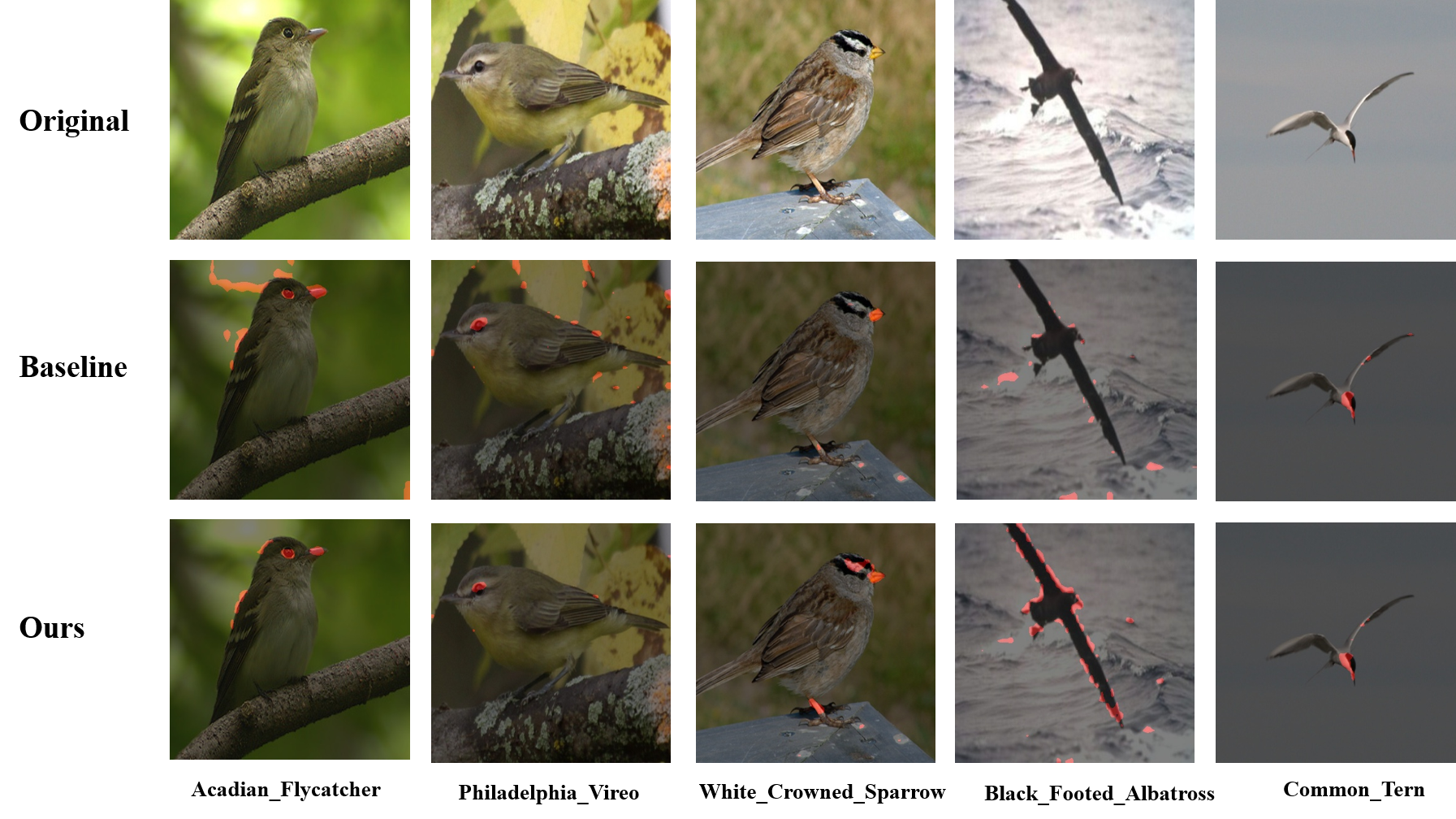}}
\caption{Visualization results on CUB-200-2011, the first row show the original image, second shows the results of Swin-T, bottom shows the results of Swin-T with PIM.}
\label{fig6}
\end{figure*}

\subsection{Ablation study}

This section explores the components in the proposed PIM that affect accuracy. We first introduce the evaluation method used, and the experimental results for different evaluation methods are presented in Table \ref{tab4}. We calculate an accuracy rate for each block and combiner's prediction results, respectively. The first five rows in Table \ref{tab4} represent these accuracy rates, and the average of the largest one to five of the prediction scores is selected as the prediction result. In Table \ref{tab4}, the first five rows represent these accuracy rates, and the average of the largest one to five of the above prediction scores is selected as the prediction result, and the results are displayed in top-1 score prediction to top-5 score prediction respectively. We can see that the average of all prediction scores achieves the best result.

\subsubsection{Number of Selections}

Swin-T has four blocks, each block outputs the number of regions selected by the Weakly Supervised Selector (num selects in \textbf{Algorithm1}), which is represented by a list in this paper. The first number in the list represents the number of selected areas for the first block, and so on.

We first discuss the number of selected areas. In this experiment, Swin-T is used as the backbone, which has four blocks. In Table \ref{tab5}, the list in the selections number represents the number of selected areas of the four blocks. The fourth column of Table \ref{tab5} shows the top-1 accuracy. It can be observed that the number of selected areas has little effect on the accuracy rate. However, the number of  selected areas has a great influence on the amount of operation of the Combiner, because the number of input nodes will directly affect the amount of parameters of the model required for subsequent mixing. Considering the trade-off between the amount of operation and the accuracy rate, we use [256, 128, 64, 32] as the final number selected areas.

\subsubsection{Selected-area and Dropped-area}

Then we discuss the impact of the selected area and the dropped area on the accuracy. We experiment with two different selection sizes, [256, 128, 64, 32] and [2048, 512, 256, 128]. Table \ref{tab6} shows that no matter how many are selected, there is a certain difference in the accuracy of the selected area and the dropped area, although the difference is not as large as imagined when the number of selected is small. If the confidence score is not enough, it can be further excluded. If threshold=0.9 is used as the boundary, it can be observed that the dropped area is still generally low.

\subsubsection{Combiner}

There are different kinds of combiners mentioned in the previous chapter. We experiment with two structures, multi layer perceptron (MLP) and graph convolutional network (GCN). The experimental results are shown in Table \ref{tab7}. In the MLP architecture, the reason we only use one layer is the limitation of the amount of parameters and the amount of calculation. If we want to use multiple layers, we must reduce the output feature size or reduce the number of selections. Because the top-1 accuracy of single-layer MLP is already lower than 0.61\% of single-layer GCN structure, we choose to use GCN as the main architecture of Combiner. Based on the experimental results shown in Table 7, we choose a single-layer GCN as the Combiner, which has the highest accuracy and less computation.

\subsubsection{Pooling Number}

We found that the number of nodes in the Combiner will affect the accuracy. We discuss the single-layer, double-layer and three-layer GCN structure. In Table \ref{tab8}, Numbers represents the ratio of the output nodes of each layer to the number of Combiner input nodes. 1/128 means that the output number of nodes in this layer is 1/128 of the number of Combiner input nodes. Table \ref{tab8} shows that the best result occurs with a one-layer structure, and the ratio of the number of output nodes to input nodes is 1/32. 

The number of selected areas can affect the accuracy by about $\pm$0.4\%, while the GCN architecture (including the number of layers and the amount of fusion) can affect the accuracy by about $\pm$0.6\%, which greatly affects the results. The experimental results show that the number of selected regions is best with the setting of [2048, 512, 128, 32] or [256, 128, 64, 32], and the best setting of Combiner is a single-layer architecture of 1/64 or 1/32.

\subsubsection{Pretrained-Effect}

In Table \ref{tab1}, the first and second row shows the recogniztion results of ResNet-50. The first row is the result in TransFG\cite{TransFG} and the second row is the result of our retrained model. The difference is that we use ImageNet-21K as the pre-training model, and it can be observed that the retrained model can effectively improve the accuracy by 3.7\%.

\subsubsection{Visualization}

Finally, we show the effect of adding each component to the backbone in Table \ref{tab9}. When FPN is added, the overall accuracy can be increased by 0.1\%, then by adding Weakly Supervised Selector, it can be increased by 0.1\%, and by adding Combiner, it can be increased by 0.7\%. This results show that the accuracy of the backbone model can be effectively improved after two feature fusions. In particular, fusion of various scale features through a GCN-Combiner can significantly improve the accuracy.

We use the above experiments to verify the capabilities of the architecture. The most important point is that it is easy to implement end-to-end training. Finally, the visualization results are used to explore whether this architecture can focus on the discriminative area, and the results are shown in Fig.\ref{fig6}.

The heat map in Fig.\ref{fig6} is calculated using Grad-CAM for the 4 block outputs of Swin-T. The top part of the figure shows the original image, the middle shows the visualization results of the baseline model, and the bottom shows the visualized heat maps after adding the architecture proposed in this paper.

It can be observed that compared to the baseline model, our method can focus more on the key points, and the background noise will be filtered out relatively cleanly. From the cases of Acdian Flycatcher and Philadelphia Vireo, in addition to focusing more on the object itself, this method can also increase the number of key positions. In the case of the white-crowned sparrow, this method also focuses on the parts around the eyes compared to the focused position of the baseline model. If the background is clean such as the case of Common Tern, the effect of the proposed method will not be significantly different from the baseline model.

\section{Conclusion}

In this paper, we have proposed a novel plug-in module that can be easily applied to popular backbone networks to learn local region features through differentiation. Experimental results show that the proposed method significantly improves the accuracy of fine-grained visual classification and outperforms state-of-the-art approaches.

\section*{Acknowledgment}

This work was financially supported by the National Taiwan Normal University (NTNU) within the framework of the Higher Education Sprout Project by the Ministry of Education(MOE) in Taiwan, sponsored by Ministry of Science and Technology, Taiwan, R.O.C. under Grant no. MOST 110-2221-E-003-026, 110-2634-F-003 -007, and 110-2634-F-003 -006. In addition, we thank to National Center for High-performance Computing (NCHC) for providing computational and storage resources.

\bibliographystyle{ieeetr}
\bibliography{refs}

\end{document}